\documentclass{article}
\usepackage{interspeech2005,amssymb,amsmath,epsfig}
\usepackage{indentfirst}
\usepackage{tipa}

\ninept
\def\reg{{\rm\ooalign{\hfil
     \raise.07ex\hbox{\scriptsize R}\hfil\crcr\mathhexbox20D}}}

\title{An elitist approach for extracting automatically well-realized speech sounds with high confidence}

\makeatletter
\def\name#1{\gdef\@name{#1\\}}
\makeatother \name{{\em Jean-Baptiste Maj, Anne Bonneau, Dominique
Fohr}\\{\em Yves Laprie}}

\address{Speech Group  \\
Loria-INRIA / CNRS, Nancy, France \\ {\small \tt jbmaj@loria.fr} }
\begin{document}
\maketitle
\begin{abstract}
This paper presents an `elitist approach' for extracting
automatically well-realized speech sounds with high confidence.
The elitist approach uses a speech recognition system based on
Hidden Markov Models (HMM). The HMM are trained on speech sounds
which are systematically well-detected in an iterative procedure.
The results show that, by using the HMM models defined in the
training phase, the speech recognizer detects reliably specific
speech sounds with a small rate of errors.
\end{abstract}

\section{Introduction}

The present article introduces a method called `elitist approach'
for extracting automatically well-realized speech sounds with high
confidence. By well-realized, we mean that the speech sounds have
well marked phonetic features. From an acoustical and perceptual
point of view, a same sound in the same phonetic context has more
or less marked acoustics cues and a highly variable level of
intelligibility. Considering that well marked phonetic features
should be identified very reliably in automatic speech recognition
we defined a series of cues, called `strong cues', specially
designed to identify this kind of feature \cite{Bonneau95}.

These cues were defined from acoustic-phonetic knowledge and
tested by means of a semi-automatic rule based speech recognition
system. Since the purpose of `strong cues' is to find out features
that are well-realized from an acoustical point of view and make
no error, they are not triggered systematically.

A very reliable detection of well-realized sounds may lead to two
kinds of application : provide an automatic speech recognition
system with reliable information on the one hand, and improve the
intelligibility of speech through the enhancement of well-realized
sounds on the other hand.

In order to detect `well-realized' sounds in a fully automatic
manner, we design in the present article, an elitist learning of
HMM that make very reliable sound models emerge. The learning is
iterated by feeding sounds identified correctly at the previous
iteration in to the learning algorithm.

The elitist approach is based on a speech recognizer and a DTW
algorithm. The speech recognition system works with Hidden Markov
Models (HMM) and a learning stage is performed to train these
models. The DTW algorithm compares the output of the speech
recognizer and the phonetic annotation of corpora for identifying
well- and wrongly-detected speech sounds. The phonetic annotations
of the corpora are changed according to the results of the DTW
algorithm, and an iterative procedure is carried out to improve
the accuracy of the HMM-based phoneme models.  In this study, the
well-detected phonemes are assumed to be well-realized speech
sounds.

A similar approach was proposed by Schwenk \cite{Schwenk99} to
guide the learning of multilayer perceptrons in the context of
automatic speech recognition. This approach is based on a
composite classifier which emphasizes certain patterns. More
recently, Chang et al. \cite{Chang01} developed an approach to
perform the articulatory labelling of a speech database through a
connexionnist method. The method removed spectral vectors which
were badly identified after a first learning phase. Thus, in a
second learning phase, the strategy created multilayer perceptrons
by only using the correctly identified spectral vectors.


\section{Corpus material}\label{corpus}
The corpus `Bref 120' for developing and evaluating speech
recognition systems is used \cite{Bref120}. In this corpus, 120
adults read sentences taken from the newspaper `Le Monde', and in
total 66553 sentences are available. Recorded speech (sampling
frequency 16kHz) and texts of the different sentences are
available. By using a phonetic annotation procedure, the phonemes
and their duration are defined and calculated for every sentence,
excepted for the sentences containing proper nouns. Hence, 56519
sentences are annotated phonetically.

In this study, Bref 120 is shared in two different corpora for
learning (A) and evaluation (B) stages. Both corpora are annotated
phonetically with two different conditions. The first does not
take into account the phonetic context and the second takes into
account the context. Without context, the corpus is annotated with
36 labels. With context, the unvoiced stops (/p,t,k/) and
fricatives (/f,s,\textesh/) are annotated as a function of the
following vowels. In total, there are six different classes of
context and the corpus is annotated with 66 labels. The classes
are defined as a function of the following vowel features.

\section{Elitist approach}
As already mentioned, the elitist approach is based on a speech
recognizer and a Dynamic Time Warping (DTW) algorithm. The speech
recognizer is involved in two phases. The first, called `training
phase', creates the HMM-based speech sound models. The HMM models
have three states, a simple left-right topology and a mixture of
64 gaussians. The grammar used is a simple phoneme loop. The
second, called `recognition phase', uses the speech recognizer and
the HMM defined in the training phase to detect speech sounds in
acoustic signals. To perform the training and the recognition
phases, the software Espere developed by the Speech Group of Loria
is used \cite{Fohr00}. In the sequel, we described the learning
and evaluation stages of the elitist approach.

\subsection{Learning stage}
\begin{figure}[!h]
    \begin{center}
        \includegraphics[width=7cm]{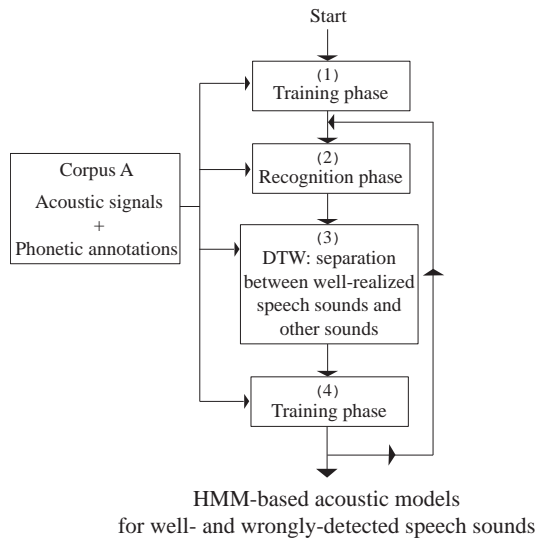}\caption{Scheme of the learing stage for the elitist approach.}\label{strategy}
    \end{center}
\end{figure}

During the learning stage, the corpus A with both conditions of
phonetic annotations is used. The goal of this stage is to train
the HMM-based phoneme models. The strategy of the elitist approach
is depicted in \textbf{figure \ref{strategy}}.

(1) A training phase is performed to create the HMM-based phoneme
models by using the acoustic signals and phonetic annotations of
the corpus A.

(2) A speech recognition is carried out with the corpus A and the
HMM models defined in stage (1).

(3) The DTW algorithm compares phonetic annotations of corpus A
and the output of the speech recognition system. The
wrongly-detected phonemes have their phonetic annotations changed.
For instance, a phoneme annotated /p/ in the corpus A and
wrongly-detected by the speech recognition system becomes /p'/.
The phonetic annotation of the corpus A is changed after each
iteration, and consequently the number of systematically
well-detected speech sounds decreases.

(4) A training phase is again performed to create new HMM-based
acoustic models of the systematically well-detected speech sounds
and other speech sounds. \\

After this stage, an iterative procedure is carried out with the
stages (2), (3) and (4) to improve the accuracy of the HMM for the
systematically well-detected speech sounds and other speech
sounds. At the first iteration, the HMM defined in stage (1) are
used in stage (2), whereas at other iterations the HMM defined in
stage (4) are used in stage (2).

To sum up, we design an elitist learning of HMM that makes very
reliable sound models emerge. The learning is iterated by feeding
sounds identified correctly at the previous iteration into the
learning algorithm. Thus, the training phase produces models for
the systematically well-detected speech sounds on the one hand,
and standard models for the other sounds on the other hand.

\subsection{Evaluation stage}
The performances of the elitist approach is evaluated with the
corpus B. For this purpose, a speech recognition is carried out on
the corpus B with HMM-based phoneme models of systematically
well-detected speech sounds and models for other speech sounds.
Afterwards, the DTW algorithm compares the output of the speech
recognizer with the phonetic annotations of the corpus. Hence, the
performance of the elitist approach can be estimated.

It is expected that the rate of trigger action for the HMM
corresponding to the well-detected speech sounds is high while the
rate of errors is small.

\section{Results}

\begin{figure}[!t]
    \centerline{\includegraphics[width=8cm]{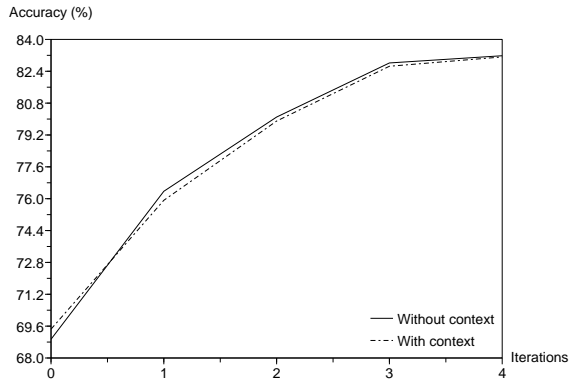}}\caption{Learning stage: the
    accuracy (\%) represents the rate of systematically
    well-detected speech sounds which are still correctly
    identified at iteration (n). The speech recognizer at iteration (n) uses the HMM-models of the systematically well-detected
    speech sounds trained with the sounds labeled as
    `well-detected' at the iteration (n-1). The corpus A is used.}\label{Figure1}
\end{figure}

\subsection{Learning stage}
During the learning stage, the elitist approach iterates on
phonemes systematically well-detected. This means that the number
of speech sounds systematically well-detected decreases at every
iteration. After a few iterations, the speech sounds
systematically well-detected are expected to have well marked
acoustic cues.

The performance of the learning phase is evaluated by the
identification rate (or accuracy) of previously well-detected
speech sounds, and the percentage of systematically well-detected
phonemes available in the corpus A. The higher the identification
rate and the percentage of items, the better the performance of
the learning phase.

At the output of the recognition phase (stage (2) of
\textbf{figure \ref{strategy}}), the accuracy of the
systematically well-detected speech sounds is measured at each
iteration. Thanks to the DTW algorithm, the numbers of
well-detected speech sounds ($Ok$), insertions ($Ins$),
substitutions ($Sub$) and omissions ($Omi$) are known. Thus, the
accuracy of the systematically well-detected speech sounds can be
calculated by:
\begin{eqnarray}\label{accuracy}
    Accuracy=\frac{Ok-Ins}{Sub+Omi+Ok}
\end{eqnarray}

When the contextual models are used, the consonants
/p,t,k,f,s,$\int$/ are labeled with the name of the following
vowels. However, for calculating the accuracy, if one element of a
class is detected as another element of the same class, then this
element is considered as well-detected ($Ok$) and not as
substituted ($Sub$).

\begin{figure}[!h]
    \centerline{\includegraphics[width=8cm]{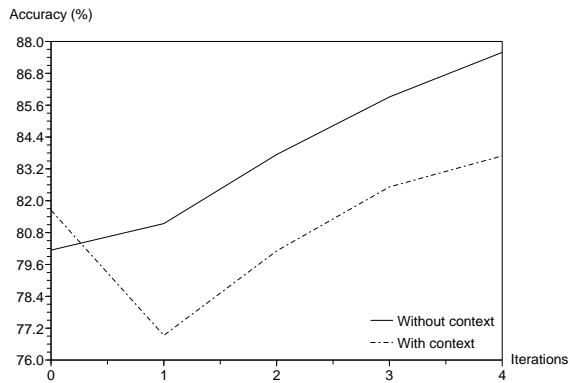}}\caption{Learning stage: accuracy (\%) of the elitist
    approach for the speech sounds /p,t,k,f,s,\textesh/ as a function of iterations. The corpus A is used.}\label{Figure2}
\end{figure}

\textbf{Figure \ref{Figure1}} and \textbf{\ref{Figure2}} show the
accuracy at the output of the recognition phase for all the
phonemes and the specific phonemes (/p,t,k,f,s,\textesh/),
respectively.

As expected at the first iteration (0), the contextual models give
a better accuracy than the non contextual models. However at the
other iterations (1, 2, 3 and 4), the contextual models give the
best accuracy.

A statistical analysis was carried out with all the phonemes to
compare both phonetic annotation conditions (\textbf{figure
\ref{Figure1}}). The analysis estimated that there are significant
differences between the contextual and non contextual models at
the iteration 0, 1, 2 and 3. At the iteration 4, the statistical
analysis estimated that they are no significant differences
between both conditions of phonetic annotations.

With the specific phonemes (/p,t,k,f,s,\textesh/), \textbf{figure
\ref{Figure2}} shows that the accuracy increases at each iteration
when the non contextual models are used. However with context, the
rate decreases significantly between the first and the second
iteration. This can be explained by the number of items which is
available in the corpus for every phoneme.

\begin{figure}[!h]
    \centerline{\includegraphics[width=8cm]{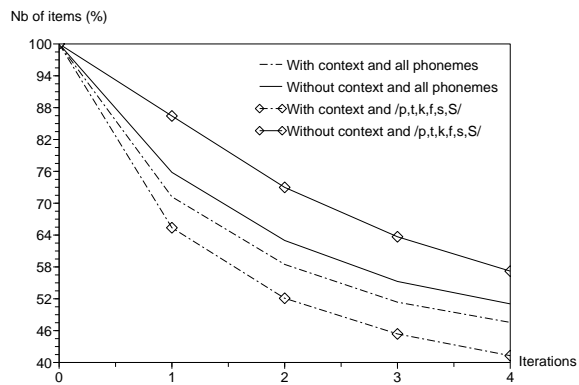}}\caption{Percentage
    of items annotated as well-detected in the corpus A.}\label{Figure3}
\end{figure}

Indeed, the performance of the speech recognizers may depend on
the number of items which are available in the corpus to train
HMM-based phoneme models. It seems that at the first iteration,
the number of items is high enough to create accurate HMM for each
phoneme. At the second iteration, the phonetic annotation is
changed and the number of systematically well-detected speech
sounds decreased significantly (\textbf{figure \ref{Figure3}}).
For a few phonemes, the number of items is around one hundred. The
small number of items seems to decrease the quality of the
HMM-based acoustic models using 64 gaussians, and then affects the
performance of the speech recognizer.

\textbf{Figure \ref{Figure3}} shows the percentage of items
annotated as well-detected after each iteration in the corpus A.
The elitist approach is more selective with context than without
context. For instance with specific phonemes /p,t,k,f,s,\textesh/
at the second iteration, 86.4\% of the phonemes are still
annotated as well-detected when the non contextual models are
used. At the same iteration but with context, 65.4\% of the
phonemes are annotated as well-detected.

To sum up, the best identification rate (83\%) of the
systematically well-detected speech sounds for all phonemes is
obtained at the fifth iteration (\textbf{figure \ref{Figure1}}).
This rate is the same for both phonetic annotation conditions. The
identification rate is measured for the systematically
well-detected speech sounds /p,t,k,f,s,\textesh/ (\textbf{figure
\ref{Figure2}}). The best accuracy (88\%) is also obtained at the
fifth iteration when the non contextual models are used. Finally,
at the fifth iteration without context, 51\% of all phonemes and
57\% of the phonemes /p,t,k,f,s,\textesh/ are systematically
well-detected (\textbf{figure \ref{Figure3}}).

Hence, the non contextual HMM models created at the fifth
iteration are used to perform the evaluation stage.

\subsection{Evaluation stage}
Speech recognition is performed with the HMM models created in the
learning phase on well identified speech sounds and other speech
sounds. The goal of the evaluation stage is to assess the rate of
trigger action for the systematically well-detected speech sounds
and the rate of false alarms.

Thus, we expect to have a significant identification rate of the
systematically well-detected speech sounds with a minimum rate of
errors. In the sequel, we are mainly interested in the
identification rate of the specific speech sounds
/p,t,k,f,s,\textesh/ in order to develop a speech enhancement
technique \cite{Colotte02}. We also assume that systematically
well-detected speech sounds by speech recognition are
well-realized sounds. Conversely, the other sounds are assumed to
be not so well-realized sounds.

Since the speech recognizer uses the HMM model of the
systematically well-detected and other speech sounds, the global
rate of correct detection of the elitist approach is 79.2\% on
average over the unvoiced speech sounds (\textbf{table
\ref{Table1}} and \textbf{\ref{Table2}}). \textbf{Table
\ref{Table1}} shows the percentage of unvoiced speech sounds
identified as well-realized /p,t,k,f,s,\textesh/, and
\textbf{table \ref{Table2}} shows the percentage of unvoiced
speech sounds identified as not so well-realized
/p',t',k',f',s',\textesh'/. Thus, the global rate of 79.2\% is
shared into a rate of 55.3\% where the unvoiced speech sounds are
identified as well-realized (\textbf{table \ref{Table1}}), and a
rate of 23.9\% where the unvoiced speech sounds are identified as
not so well-realized (\textbf{table \ref{Table2}}).

The rate of false alarms is rather small for the unvoiced stops
and fricatives considered as well-realized (\textbf{table
\ref{Table1}}). For instance, 1.63\% of the speech sounds /t/ are
identified as /p/ by the elitist approach. The rate of false
alarms is 0.82\% on average.

The rate of false alarms within the not so well-realized unvoiced
speech sounds is rather small and is about 0.7\% (\textbf{table
\ref{Table2}}).

\begin{table}[!t]
    \begin{center}
        \includegraphics[width=8cm]{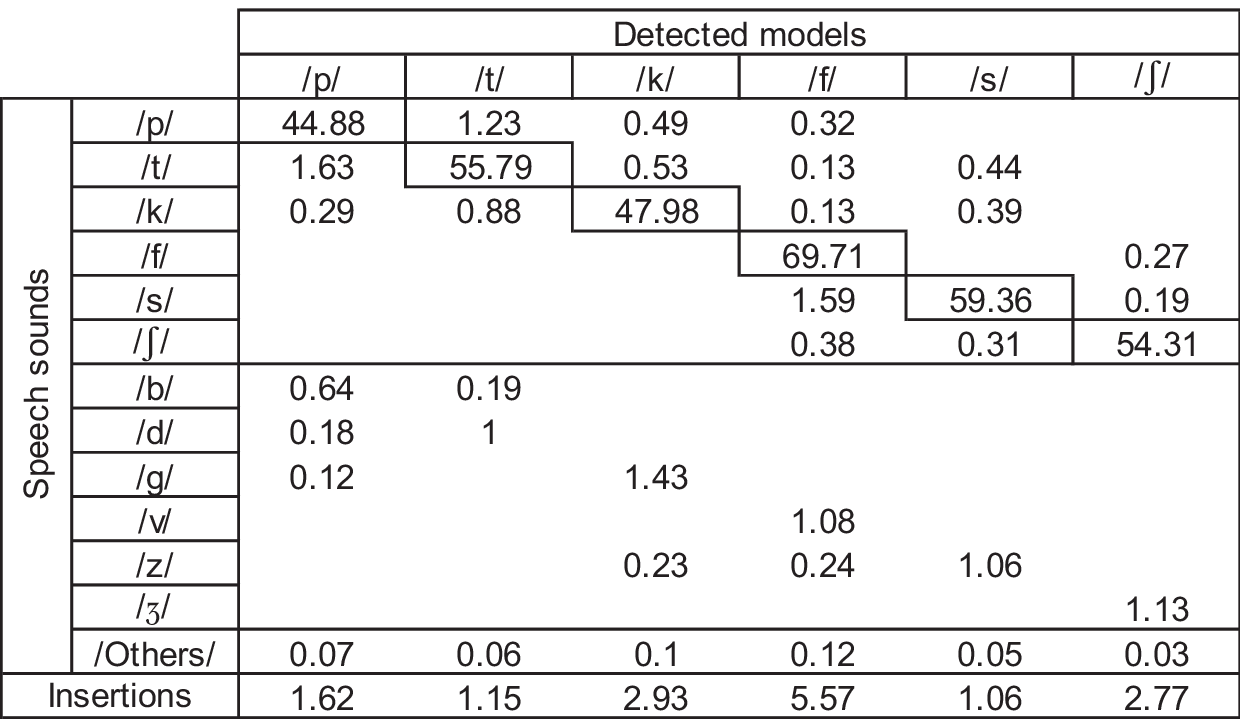}
    \end{center}
    \caption{Rate (\%) of trigger action for the HMM-based models of the well-realized unvoiced speech sounds.
    The corpus B is used. Only the percentages above 0.01\% are mentioned.}\label{Table1}
\end{table}


\begin{table}[!h]
    \begin{center}
        \includegraphics[width=8cm]{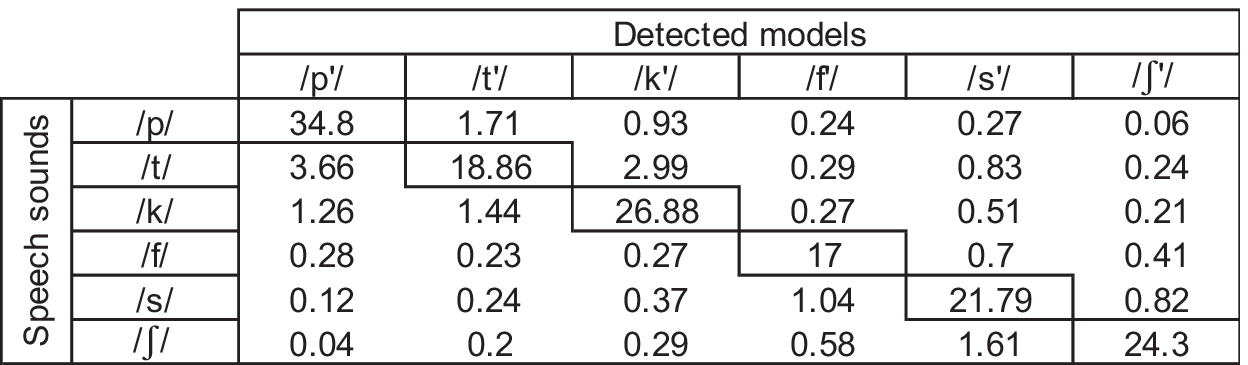}
    \end{center}
    \caption{Rate (\%) of trigger action for the HMM-based models of the not so well-realized unvoiced speech sounds.
    The corpus B is used. Only the percentages above 0.01\% are mentioned.}\label{Table2}
\end{table}

To sum up, the speech recognizer identifies 55\% of the speech
sounds as well-realized, with a small rate of errors. Hence, the
elitist approach can extract automatically speech sounds, assumed
to be well-realized, with high confidence. With the aim of
developing a speech enhancement technique, this is an important
result. Indeed, we will be able to perform modifications with high
confidence on the speech sounds detected by the speech recognizer
using the HMM models of the systematically well-detected speech
sounds of the learning phase.

\section{Perspectives}
We have assumed that the systematically well-detected speech
sounds have well marked phonetic features. This is why we have
created HMM-based acoustic models for the supposed well-realized
speech sounds. Our next work is to prove that the well-detected
speech sounds have well marked phonetic features. To perform this
work, a system devoted for the identification of stop consonants
will be used \cite{Bonneau95}.

This system exploits acoustic detectors designed in the study of
Bonneau et al. to evaluate acoustic cues provided by burst and
formant transitions. Since spectral characteristics of burst
release are more relevant to identify stop consonants, formants
are tracked automatically and a segmentation algorithm separates
the burst release from the frication noise. Spectral cues,
particularly the emergence and the frequency of the most prominent
peak, turned out to be very efficient to detect strong cues.

The work will consist in studying the spectral cues of the
well-realized speech sounds.



\section{Conclusions}
Promising results were obtained with the elitist approach for
extracting automatically unvoiced stops and fricatives. On
average, 55\% of the specific speech sounds are classified as
well-realized speech sounds with a small percentage of errors.

A next work will consist in studying the acoustic cues of the
speech sounds assumed as well-realized and not so well-realized
speech sounds. Finally, perceptual tests will be carried out for
testing if the amplification of the well-detected unvoiced stops
and fricatives can improve speech intelligibility.

\section{Acknowledgement}
Partly funded by Voice Web Initiative - HP Philantropy and
Education in Europe.
\bibliographystyle{IEEEtran}

\end{document}